# Let's Get Ready to Rumble:
# Crossover Versus Mutation Head to Head

**Kumara Sastry**
**David E. Goldberg**

# Let's Get Ready to Rumble: Crossover Versus Mutation Head to Head

Kumara Sastry[1,2], David E. Goldberg[1,3]
[1]Illinois Genetic Algorithms Laboratory (IlliGAL)
[2]Department of Material Science and Engineering
[3]Department of General Engineering
University of Illinois at Urbana-Champaign
Urbana, IL 61801
{ksastry,deg}@uiuc.edu

**Abstract**

This paper analyzes the relative advantages between crossover and mutation on a class of deterministic and stochastic additively separable problems. This study assumes that the recombination and mutation operators have the knowledge of the building blocks (BBs) and effectively exchange or search among competing BBs. Facetwise models of convergence time and population sizing have been used to determine the scalability of each algorithm. The analysis shows that for additively separable deterministic problems, the BB-wise mutation is more efficient than crossover, while the crossover outperforms the mutation on additively separable problems perturbed with additive Gaussian noise. The results show that the speed-up of using BB-wise mutation on deterministic problems is $\mathcal{O}(\sqrt{k}\log m)$, where $k$ is the BB size, and $m$ is the number of BBs. Likewise, the speed-up of using crossover on stochastic problems with fixed noise variance is $\mathcal{O}(m\sqrt{k}/\log m)$.

## 1 Introduction

Great debate between crossover and mutation has consumed much ink and many trees over the years. When mutation works it is lightening quick and uses small or non-extent populations. Crossover when it works, seems to be able to tackle more complex problems, but getting the population size and other parameters set is a challenge. Comparisons between the two are usually written by a researcher with an axe to grind. Comparisons are usually empirical, the basis for comparison is implicitly or explicitly unfair, and theory is non-existent. Wouldn't it be nice to compare our two favorite genetic operators on a fair basis in an interesting class of problems and let them slug it out head to head.

That's what we do here. Assuming that both the recombination and mutation operators possess linkage (or neighborhood) knowledge, we pit them against each other for solving boundedly difficult additively separable problems with and without the presence of additive exogenous noise. We use a recombination operator that exchanges building blocks (BBs) without disrupting them and a mutation operator that performs local search among competing building-block neighborhood. The motivation for this study also comes from recent local-search literature, where authors have highlighted the importance of using a good *neighborhood* operator (Barnes, Dimova, & Dokov, 2003;

Watson, 2003). However, a systematic method of designing a good neighborhood operator for a class of search problems is still an open question. We investigate whether using a neighborhood operator that searches among competing BBs of a problem would be advantageous and if so under what circumstances.

This paper is organized as follows. The next section gives a brief review of related literature. We provide an outline of the crossover-based and mutation-based genetic algorithms (GAs) in Section 3. Facetwise models are developed to determine the scalability of the crossover and the BB-wise mutation-based GAs for deterministic fitness functions in Section 4 and for noisy fitness functions in Section 5. Finally, we discuss future research directions followed by conclusions.

## 2 Literature Review

Over the last few decades many researchers have empirically and theoretically studied where genetic algorithms excel. An exhaustive literature review is out of the scope of this paper, and therefore we present a brief review of related theoretical studies.

Several authors have analyzed the scalability of a mutation based hillclimber and compared it to scalability of different forms of genetic algorithms, such as breeder genetic algorithm (Mühlenbein, 1991; Mühlenbein, 1992), an ideal genetic algorithm (Mitchell, Holland, & Forrest, 1994), and a genetic algorithm with *culling* (Baum, Boneh, & Garrett, 2001). Goldberg (Goldberg, 1999) gave a theoretical analysis of deciding between a single run with a large population GA and multiple runs with several small population GAs, under the constraint of fixed computational cost. He showed that for uniformly-scaled problems a single run of large population GA was advantageous, while for exponentially-scaled problems small population GAs with multiple restarts were better. Srivastava and Goldberg (Srivastava & Goldberg, 2001; Srivastava, 2002) empirically verified and analytically enhanced the *time-continuation* theory put forth by Goldberg (Goldberg, 1999). Recently, Cantú-Paz and Goldberg (Cantú-Paz & Goldberg, 2003) investigated scenarios under which multiple runs of a GA are better than a single GA run. For an exhaustive review of studies on the advantages/disadvantages of multiple populations both under serial and parallel GAs over a single large-population GA, the reader is referred elsewhere (Cantú-Paz, 2000; Srivastava, 2002; Luke, 2001; Fuchs, 1999) and to the references therein.

While many of the related studies (Goldberg, 1999; Srivastava & Goldberg, 2001; Cantú-Paz & Goldberg, 2003) assumed fixed genetic operators, with no knowledge of building-block structure, in this paper, we assume that the recombination and mutation operators have linkage (or neighborhood) knowledge. While the linkage information is usually unknown for a given search problem, a variety of linkage identification methods can be used to design the operators (see Goldberg (Goldberg, 2002), Sastry and Goldberg (Sastry & Goldberg, 2004), and references therein).

## 3 Preliminaries

The objective of this paper is to predict the relative computational costs of a crossover and an ideal-mutation based algorithm for additively separable problems with and without additive Gaussian noise. Before developing models for estimating the computational costs, we briefly describe the algorithms and the assumptions used in the paper.

## 3.1 Selectorecombinative Genetic Algorithms

We consider a generationwise selectorecombinative GA with non-overlapping populations of fixed size (Holland, 1975; Goldberg, 1989). We apply crossover with a probability of 1.0 and do not use any mutation. We assume binary strings of fixed length as the chromosomes. To ease the analytical burden, the selection mechanism assumed throughout the analysis is binary tournament selection (Goldberg, Korb, & Deb, 1989). However, the results can be extended to other tournament sizes and other selection methods in a straightforward manner. The recombination method used in the analysis is a uniform building-block-wise crossover (Thierens & Goldberg, 1994). In uniform BB-wise crossover, two parents are randomly selected from the mating pool and their building blocks in each partition are exchanged with a probability of 0.5. Therefore, none of the building blocks are disrupted during a recombination event. The offspring created through crossover entirely replace the parental individuals.

## 3.2 Building-Block-Wise Mutation Algorithm (BBMA)

In this paper we consider an *enumerative BB-wise mutation* operator, in which we start with a random individual and evaluate all possible schemas in a given partition. That is, for a building-block of size $k$, we evaluate all $2^k$ individuals. The best out of $2^k$ individuals is chosen as a candidate for mutating BBs of other partitions. In other words, the BBs in different partitions are mutated in a sequential manner. For a problem with $m$ BBs of size $k$ each, the BBMA can be described as follows:

1. Start with a random individual and evaluate it.

2. Consider the first non-mutated BB. Here the BB order is chosen arbitrarily from left-to-right, however, different schemes can be—or may required to be—chosen to decide the order of BBs.

3. Create $2^k - 1$ unique individuals with all possible schemata in the chosen BB partition. Note that the schemata in other partitions are the same as the original individual (from step 2).

4. Evaluate all $2^k - 1$ individuals and retain the best for mutation of BBs in other partitions.

5. Repeat steps 2–4 till BBs of all the partitions have been mutated.

We use an enumerative BB-wise mutation for simplifying the analysis and a greedy BB-wise method can improve the performance of the mutation-based algorithm. A straightforward Markov process analysis—along the lines of (Mühlenbein, 1991; Mühlenbein, 1992)—of a greedy BB-wise mutation algorithm indeed shows that the greedy method is on an average better than the enumerative one. However, the analysis also shows that differences between the greedy and enumerative BB-wise mutation approaches are little, especially for moderate-to-large problems. Moreover, the computational costs of an enumerative BB-wise mutation bounds the costs of a greedy BB-wise mutation.

# 4 Crossover vs. Mutation: Deterministic Fitness Functions

In this section we analyze the relative computational costs of using a selectorecombinative GA or a BB-wise mutation algorithm for successfully solving deterministic problems of bounded difficulty.

The objective of the analysis is to answer whether a population-based selectorecombinative GA is computationally advantageous over a BB-wise-mutation based algorithm. If one algorithm is better than the other, we are also interested in estimating the savings in computational time. Note that unlike earlier studies, we assume that the building-block structure is known to both the crossover and mutation operators.

We begin our analysis with the scalability of selectorecombinative genetic algorithms followed by the scalability of the BB-wise mutation algorithm.

## 4.1 Scalability of Selectorecombinative GA

Two key factors for predicting the scalability and estimating the computational costs of a genetic algorithm are the convergence time and population sizing. Therefore, in the following subsections we present facetwise models of convergence time and population sizing.

### 4.1.1 Population-Sizing Model

Goldberg, Deb, & Clark (Goldberg, Deb, & Clark, 1992) proposed population-sizing models for correctly deciding between competing BBs. They incorporated noise arising from other partitions into their model. However, they assumed that if wrong BBs were chosen in the first generation, the GAs would be unable to recover from the error. Harik, Cantú-Paz, Goldberg, and Miller (Harik, Cantú-Paz, Goldberg, & Miller, 1999) refined the above model by incorporating cumulative effects of decision making over time rather than in first generation only. Harik et al. (Harik, Cantú-Paz, Goldberg, & Miller, 1999) modeled the decision making between competing BBs as a gambler's ruin problem. Here we use an approximate form of the gambler's ruin population-sizing model (Harik, Cantú-Paz, Goldberg, & Miller, 1999):

$$n = \frac{\sqrt{\pi}}{2} \frac{\sigma_{BB}}{d} 2^k \sqrt{m} \log m, \tag{1}$$

where $k$ is the BB size, $m$ is the number of BBs, $d$ is the size signal between the competing BBs, and $\sigma_{BB}$ is the fitness variance of a building block. building blocks. The above equation assumes a failure probability, $\alpha = 1/m$.

### 4.1.2 Convergence-Time Model

Mühlenbein and Schlierkamp-Voosen (Mühlenbein & Schlierkamp-Voosen, 1993) derived a convergence-time model for the breeder GA using the notion of *selection intensity* (Bulmer, 1985) from population genetics. Thierens and Goldberg (Thierens & Goldberg, 1994) derived convergence-time models for different selections schemes including binary tournament selection. Bäck (Bäck, 1994) derived estimates of selection intensity for $s$-wise tournament and $(\mu, \lambda)$ selection. Miller and Goldberg (Miller & Goldberg, 1995) developed convergence-time models for $s$-wise tournament selection and incorporated the effects of external noise. Bäck (Bäck, 1995) developed convergence-time models for $(\mu, \lambda)$ selection. Even though the selection-intensity-based convergence-time models were developed for the OneMax problem, Miller and Goldberg (Miller, 1997) observed that they are generally applicable to additively decomposable problems of bounded order. Here, we use the convergence-time model of Miller and Goldberg (Miller & Goldberg, 1995):

$$t_c = \frac{\pi}{2I} \sqrt{\ell}, \tag{2}$$

where $I$ is the selection intensity, and $\ell = mk$ is the string length. For binary tournament selection, $I = 1/\sqrt{\pi}$.

Using equations 1 and 2, we can now predict the scalability, or the number of function evaluations required for successful convergence, of GAs as follows:

$$n_{\text{fe,GA}} = n \cdot t_c = \frac{\pi^2}{4} \frac{\sigma_{BB}}{d} \sqrt{k} \log m \cdot 2^k \cdot m. \tag{3}$$

### 4.2 Scalability of BB-wise Mutation Algorithm

Since the initial point is evaluated once and after that for each of the $m$ BBs, $2^k - 1$ individuals are evaluated, the total number of function evaluations required for the BBMA is

$$n_{\text{fe,BBMA}} = \left(2^k - 1\right) m + 1. \tag{4}$$

The results from the above subsections (Equations 3 and 4) indicate that while the scalability of a selectorecombinative GA is $\mathcal{O}\left(2^k m \log m\right)$, the scalability of the BBMA is $\mathcal{O}\left(2^k m\right)$. This is in contrast to a random-walk mutation algorithm with no BB knowledge which scales as $\mathcal{O}\left(m^k \log m\right)$ (Mühlenbein, 1992). By searching among building-block neighborhoods, the selectomutative algorithm scales-up significantly better than a mutation operator with no linkage information and provides a savings of $\mathcal{O}(\sqrt{k} \log m)$ evaluations over the GA. The savings comes from the extra evaluation required for the convergence and decision-making in the selectorecombinative GAs.

The speed-up—which is defined as the ratio of number of function evaluations required by a GA to that required by BBMA—obtained by using a BB-wise mutation algorithm over a selectorecombinative GA is given by

$$\eta = \frac{n_{\text{fe,GA}}}{n_{\text{fe,BBMA}}} = \mathcal{O}\left(\sqrt{k} \log m\right). \tag{5}$$

In particular, the speed-up for the OneMax problem ($k = 1$) is given by

$$\eta_{\text{OneMax}} = \frac{\frac{\pi^2}{4} m \log m}{m + 1} \approx \frac{\pi^2}{4} \log m, \tag{6}$$

and for the GA-hard m k-Trap function (Goldberg, 1987), the speed-up is given by

$$\eta_{\text{Trap}} = \frac{\frac{\pi^2}{4} \frac{\sigma_{BB}}{d} \sqrt{k} 2^k m \log m}{(2^k - 1)m + 1} \approx \frac{\pi^2}{4} \frac{\sigma_{BB}}{d} \sqrt{k} \log m. \tag{7}$$

The speed-up predicted by Equations 6 and 7 are verified with empirical results in Figures 1(a) and 1(b), respectively. The results are averaged over 900 independent runs. The results show that there is a good agreement between the predicted and observed speed-up. The results show that for deterministic additively separable problems, a BB-wise mutation algorithm is about $\mathcal{O}(\sqrt{k} m)$ times faster than a selectorecombinative genetic algorithm.

## 5 Crossover vs. Mutation: Noisy Fitness Functions

In the previous section, we observed that BB-wise mutation scales-up better than a crossover on deterministic additively separable problems. Furthermore, a selectomutative algorithm was able to overcome *deception*, one of the key factors influencing problem difficulty, using linkage (neighborhood) information and enumeration within the neighborhood. In this section we introduce another dimension of problem difficulty in *extra-BB noise* (Goldberg, 2002) and analyze if the BB-wise mutation maintains its edge over crossover. That is, we analyze whether a selectorecombinative or a selectomutative GA works better on additively separable problems with additive external Gaussian noise.

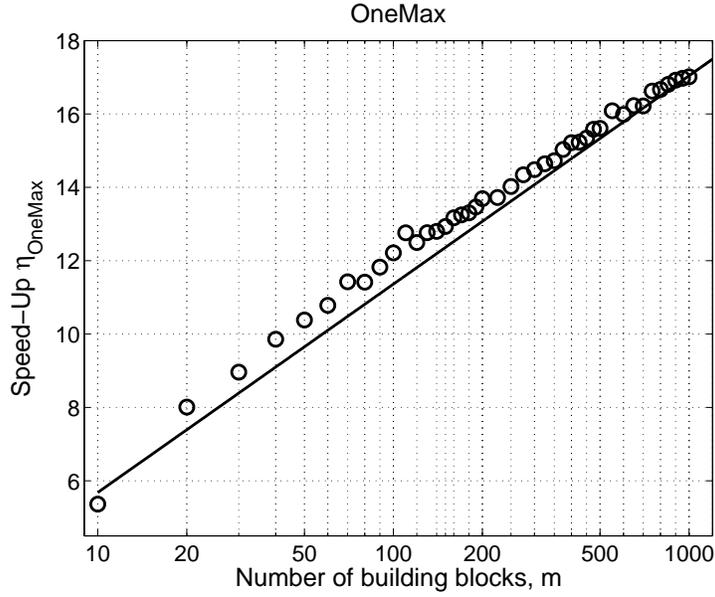

(a) OneMax

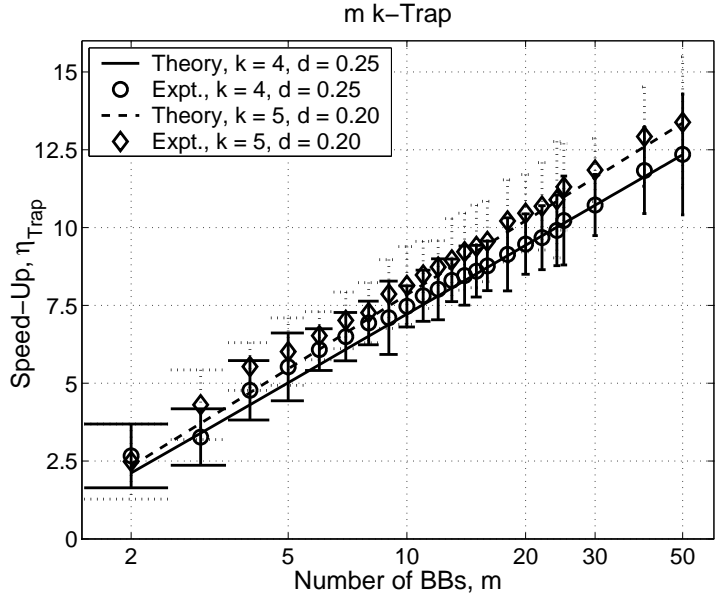

(b) m k-Trap

Figure 1: Empirical verification of the speed-up predicted for using BB-wise mutation over a selectorecombinative GA by Equations 6 and 7. The empirical results are averaged over 900 independent runs. The results show that the speed-up obtained by BB-wise mutation algorithm over a GA is $\mathcal{O}(\sqrt{k}\log m)$.

We follow the same approach outlined in the previous section and consider the scalability of crossover and mutation.

## 5.1 Scalability of Selectorecombinative GAs

Again we use the convergence-time and population-sizing models to determine the scalability of GAs under the presence of unbiased Gaussian noise. We use an approximate form of the gambler's ruin population-sizing model for noisy environments:

$$n = \frac{\sqrt{\pi}}{2} \frac{\sigma_{BB}}{d} 2^k \sqrt{m} \log m \sqrt{\left(1 + \frac{\sigma_N^2}{\sigma_f^2}\right)}, \tag{8}$$

where $\sigma_N^2$ is the variance of the noise, and $\sigma_f^2$ is the fitness variance.

We use an approximate form of Miller and Goldberg's (Miller & Goldberg, 1995) convergence-time model:

$$t_c = \frac{\pi}{2I} \sqrt{m} \sqrt{1 + \frac{\sigma_N^2}{\sigma_f^2}}. \tag{9}$$

A detailed derivation of the above equation and other approximations are given elsewhere (Goldberg, 2002; Sastry, 2001).

The population-sizing and convergence-time models indicate that the exogenous noise increases the population size and elongates the convergence time. Using equations 1 and 2, we can now predict the scalability, or the number of function evaluations required for successful convergence, of GAs as follows:

$$n_{\text{fe,GA}} = \frac{\pi^2}{4} \frac{\sigma_{BB}}{d} \sqrt{k} \log m \cdot \left(1 + \frac{\sigma_N^2}{\sigma_f^2}\right) \cdot 2^k \cdot m. \tag{10}$$

## 5.2 Scalability of BB-wise Mutation Algorithm

Unlike the deterministic case where a BB was perturbed and evaluated once, in the presence of exogenous noise we cannot rely on only a single evaluation. In other words, in the presence of noise, an average of multiple samples of the fitness should be used in deciding between competing building blocks. Now the question remains as to exactly how many samples have to be considered. This issue of exact samples of fitness required to correctly decide between competing building blocks in the presence of noise has been addressed elsewhere (Goldberg, Deb, & Clark, 1992):

$$n_s = 2c\sigma_N^2, \tag{11}$$

where $n_s$ is the number of independent fitness samples, and $c$ is the square of the ordinate of a one-sided standard Gaussian deviate at a specified error probability $\alpha$. For low error values, $c$ can be obtained by the usual approximation for the tail of a Gaussian distribution: $\alpha \approx \exp(-c/2)/(\sqrt{2c})$. In this paper we have used $\alpha = 1/m$. Equation 11 is empirically verified for the Noisy-OneMax problem in Figure 2. The results show a good agreement between the model and experiments.

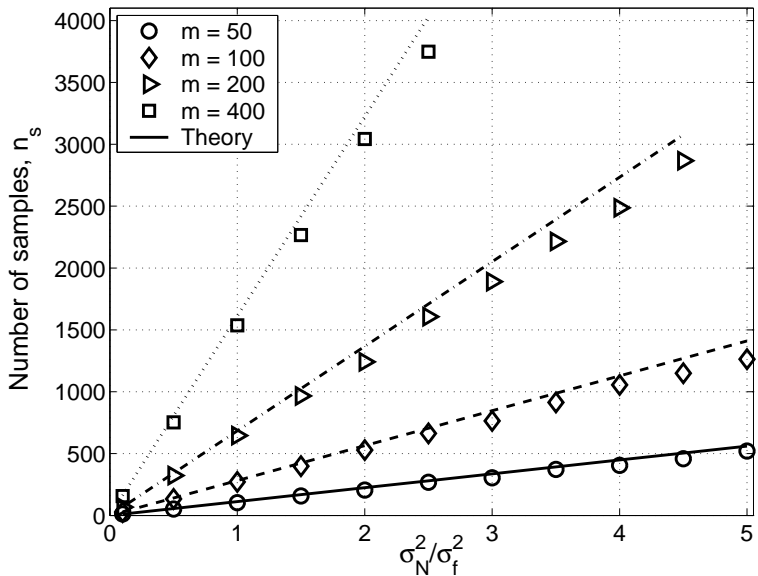

Figure 2: Comparison of the number of samples of fitness evaluations per individual required to correctly decide between competing building blocks as predicted by Equation 11 with empirical results.

Since the initial point is evaluated $n_s$ times and after that for each of the $m$ BBs, $2^k - 1$ individuals are evaluated $n_s$ times, the total number of function evaluations required for the BBMA for noisy fitness functions is given by

$$\begin{aligned} n_{\text{fe,BBMA}} &= n_s \left[ \left(2^k - 1\right) m + 1 \right], \\ &= \left( 2c \frac{\sigma_N^2}{\sigma_f^2} \cdot m \sigma_{BB} \right) \left[ \left(2^k - 1\right) m + 1 \right]. \end{aligned} \quad (12)$$

The results from the above subsections (Equations 10 and 12) indicate that under the presence of exogenous noise, a selectorecombinative GA scales as $\mathcal{O}\left(2^k m \log m(1 + \sigma_N^2/\sigma_f^2)\right)$. On the other hand, the BB-wise mutation scales as $\mathcal{O}\left(2^k m^2 (\sigma_N^2/\sigma_f^2)\right)$. Therefore, for constant values of $\sigma_N^2/\sigma_f^2$, a selectorecombinative GA is $\mathcal{O}(\sqrt{k}m/\log m)$ times faster than the BB-wise mutation. By implicitly averaging out the exogenous noise, crossover is able to overcome the extra effort needed for the convergence and decision-making. On the other hand the explicit averaging via multiple fitness samples by the BB-wise mutation leads to an order of magnitude increase in the number of function evaluations.

The speed-up—which is defined as the ratio of number of function evaluations required by mutation to that required by crossover—obtained by using a selectorecombinative over selectomutative

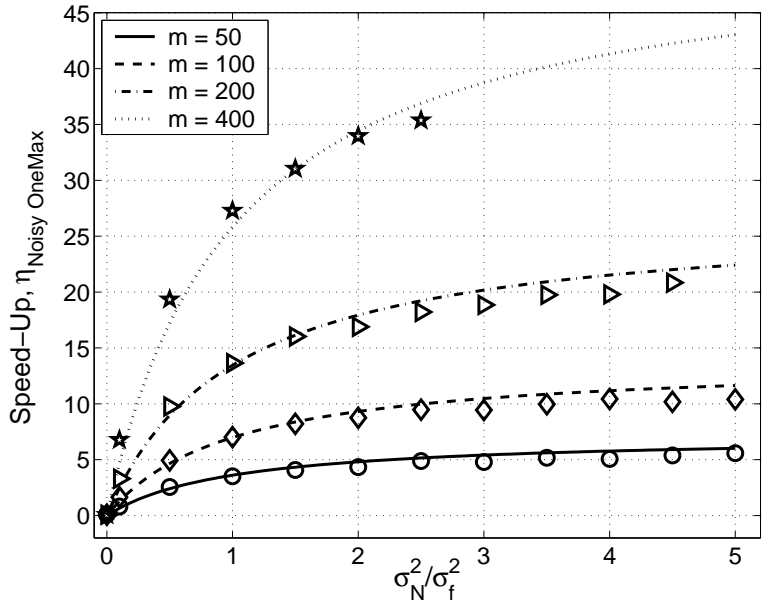

Figure 3: Empirical verification of the speed-up predicted for using BB-wise mutation over a selectorecombinative GA by Equation 14 for the OneMax problem with exogenous noise. The empirical results are averaged over 900 independent runs. The results show that a selectorecombinative GA uses significantly less number of function evaluations than the BB-wise mutation algorithm.

GA is given by

$$\eta_{\text{Noise}} = \frac{n_{\text{fe,BBMA}}}{n_{\text{fe,GA}}} = \mathcal{O}\left[\sqrt{k}\frac{m}{\log m}\left(\frac{\frac{\sigma_N^2}{\sigma_f^2}}{1+\frac{\sigma_N^2}{\sigma_f^2}}\right)\right]. \qquad (13)$$

In particular, the speed-up for the OneMax problem ($k=1$) is given by

$$\eta_{\text{Noisy OneMax}} = \frac{4c}{\pi^2}\frac{m}{\log m}\left(\frac{\frac{\sigma_N^2}{\sigma_f^2}}{1+\frac{\sigma_N^2}{\sigma_f^2}}\right). \qquad (14)$$

The speed-up predicted by Equation 14 is verified with empirical results in Figure 3. The results are averaged over 900 independent runs. The results show that there is a good agreement between the predicted and observed speed-up. The results show that for stochastic additively separable problems with constant noise variance, a selectorecombinative GA is about $\mathcal{O}(\sqrt{k}m/\log m)$ times faster than the BB-wise mutation algorithm.

## 6 Future Work

The results of this paper indicate that there are significant advantages of using a mutation operator that performs hillclimbing in the BB space and indicates many avenues of future research some of which are listed in the following:

- **Hybridization of crossover and BB-wise mutation:** While this paper consider a bounding case of crossover vs. mutation, it might be (more likely it is) more effective to use an efficient hybrid of crossover *and* mutation.

- **Designing BB-wise Mutation:** In this paper we assumed that the BB information was known, which generally is not the case. Over the last few years, effective recombination operators that adapt linkage have been developed in a systematic manner (Goldberg, 2002). On the other hand, most mutation operators, including adaptive ones, search in the local neighborhood of a solution. Furthermore, there has been growing evidence of the importance of using good neighborhood operators in determining the effectiveness of local-search methods (Barnes, Dimova, & Dokov, 2003; Watson, 2003). Despite the importance of having good neighborhood information, a general methodology for designing operators with good neighborhood information is non-existent. That is, little attention has been paid to systematically design effective mutation operators that performs local search in the building-block space (Sastry & Goldberg, 2004). The results of this paper indicate that the dividends obtained by designing BB-wise mutation operators that adaptively identify and utilize good neighborhood information can be significant.

- **Problems with overlapping building blocks:** While this paper considered problems with non-overlapping building blocks, many problems have different building blocks that share common components. An analysis similar to the one presented in this paper can be performed to predict which of the two algorithms excel. However, since the effect of overlapping variable interactions is similar to that of exogenous noise (Goldberg, 2002), based on the results of this paper crossover is likely to be more useful than the mutation for solving problems with overlapping building blocks.

- **Hierarchical problems:** One of the important class of nearly decomposable problems is hierarchical problems, in which the building-block interactions are present at more than a single level. Further investigation is necessary to analyze if BB-wise mutation can help speed-up the scalability of selectorecombinative GAs.

# 7 Summary & Conclusions

In this paper, we have introduced a building-block-wise mutation operator which efficiently searches among the competing building block (BB) neighborhood. We also compared the computational costs BB-wise mutation algorithm with a selectorecombinative genetic algorithm for both deterministic and stochastic additively separable problems. Our results show that while the BB-wise mutation provides significant advantage over crossover for deterministic problems, crossover maintains significant edge over the BB-wise mutation on stochastic problems. The results show that the speed-up of using BB-wise mutation on deterministic problems is $\mathcal{O}(\sqrt{k}\log m)$, where $k$ is the BB size, and $m$ is the number of BBs. Likewise, the speed-up of using crossover on stochastic problems with fixed noise variance is $\mathcal{O}(m\sqrt{k}/\log m)$.

## Acknowledgments

This work was sponsored by the Air Force Office of Scientific Research, Air Force Materiel Command, USAF, under grant F49620-00-0163 and F49620-03-1-0129, the National Science Foundation under grant DMI-9908252, ITR grant DMR-99-76550 (at Materials Computation Center), and ITR grant DMR-0121695 (at CPSD), and the Dept. of Energy through the Fredrick Seitz MRL (grant DEFG02-91ER45439) at UIUC. The U.S. Government is authorized to reproduce and distribute reprints for government purposes notwithstanding any copyright notation thereon.

The views and conclusions contained herein are those of the authors and should not be interpreted as necessarily representing the official policies or endorsements, either expressed or implied, of the Air Force Office of Scientific Research, the National Science Foundation, or the U.S. Government.